\documentclass[sigconf]{acmart}

\usepackage{wrapfig}
\usepackage{float}

\usepackage{caption}
\usepackage[separate-uncertainty=true]{siunitx}
\usepackage{mathtools}
\usepackage{graphicx}
\usepackage{courier}
\usepackage{epstopdf}
\usepackage{color}
\usepackage{csquotes}

\usepackage{amsmath}
\usepackage{amsfonts}
\usepackage{amssymb}
\usepackage[subnum]{cases}
\usepackage{booktabs} 
\usepackage{pifont}

\usepackage{algorithm,algorithmicx,algpseudocode}

\usepackage{amssymb}
\usepackage{pifont}

\usepackage {tikz}
\usetikzlibrary {positioning}
\definecolor {processblue}{cmyk}{0.96,0,0,0}
\usepackage{lipsum,adjustbox}

\usepackage[subnum]{cases}
\usepackage{color,soul}

\usepackage{subcaption}

\usepackage{footnote}
\makesavenoteenv{tabular} 

\usepackage{multirow}

\usepackage{booktabs,subcaption,amsfonts,dcolumn}
\newcolumntype{d}[1]{D..{#1}}

\AtBeginDocument{%
  \providecommand\BibTeX{{%
    \normalfont B\kern-0.5em{\scshape i\kern-0.25em b}\kern-0.8em\TeX}}}

\setcopyright{acmcopyright}
\copyrightyear{2018}
\acmYear{2018}
\acmDOI{10.1145/1122445.1122456}

\acmConference[XXXX '20]{XXXX '20: ACM XXXX XXXX}{2020}{XXXX, XXXX}
\acmBooktitle{XXXX '20: ACM XXXX}
\acmPrice{15.00}
\acmISBN{978-1-4503-XXXX-X/18/06}

\begin{document}

\title{Unsupervised Graph Representation by Periphery and Hierarchical Information Maximization}


\author{Sambaran Bandyopadhyay}
\affiliation{%
  \institution{IBM Research \& IISc, Bangalore}}
\email{samb.bandyo@gmail.com}

\author{Manasvi Aggarwal}
\affiliation{%
  \institution{Indian Institute of Science, Bangalore}}
\email{manasvia@iisc.ac.in}

\author{M. Narasimha Murty}
\affiliation{%
  \institution{Indian Institute of Science, Bangalore}}
\email{mnm@iisc.ac.in}






\begin{abstract}
  Deep representation learning on non-Euclidean data types, such as graphs, has gained significant attention in recent years. Invent of graph neural networks has improved the state-of-the-art for both node and the entire graph representation in a vector space. However, for the entire graph representation, most of the existing graph neural networks are trained on a graph classification loss in a supervised way. But obtaining labels of a large number of graphs is expensive for real world applications. Thus, we aim to propose an unsupervised graph neural network to generate a vector representation of an entire graph in this paper. For this purpose, we combine the idea of hierarchical graph neural networks and mutual information maximization into a single framework. We also propose and use the concept of periphery representation of a graph and show its usefulness in the proposed algorithm which is referred as \textbf{GraPHmax}. We conduct thorough experiments on several real-world graph datasets and compare the performance of GraPHmax with a diverse set of both supervised and unsupervised baseline algorithms. Experimental results show that we are able to improve the state-of-the-art for multiple graph level tasks on several real-world datasets, while remain competitive on the others.
\end{abstract}




\maketitle

\section{Introduction}\label{sec:intro}
Graph is an important data structure used to represent relational objects from multiple domains such as social networks, molecular structures, protein-protein interaction networks \cite{newman2003structure,krogan2006global} etc. Recent advent of deep learning has significantly influenced the field of network representation learning, i.e., representing a node or an edge of a graph, or the entire graph itself in a vector space \cite{wu2019comprehensive}. Graph neural networks have been shown to be powerful tools, which directly take the graph structure as input to a neural network and learn the representation by minimizing different types of supervised \cite{defferrard2016convolutional,gilmer2017neural}, semi-supervised \cite{kipf2016semi,you2019position} or unsupervised objective functions \cite{kipf2016variational,zhang2018link}. For a graph level task such as graph classification \cite{xu2018how,duvenaud2015convolutional}, GNNs jointly derive the node embeddings and use different pooling mechanisms \cite{ying2018hierarchical,lee2019self} to obtain a representation of the entire graph.

Initial works of network representation learning \cite{perozzi2014deepwalk,grover2016node2vec} focus on obtaining unsupervised vector representation of nodes of a graph, and was heavily influence by the development in the field of natural language understanding \cite{mikolov2013distributed}. Graph neural networks (GNNs) \cite{kipf2016semi} show more promising results to generate node representation in a semi-supervised fashion. However, the main challenge to obtain a graph level representation from a set of node embeddings was to develop an efficient pooling mechanism. Simple pooling technique such as taking the average of node embeddings in the final layer of a GCN \cite{duvenaud2015convolutional} and more advanced deep learning architectures that operate over the sets \cite{gilmer2017neural} are proposed. Recently, hierarchical pooling strategies \cite{ying2018hierarchical,morris2019weisfeiler} are proposed, which hierarchically convert a graph to a set of sub-communities and then to communities etc. to generate the graph representation. Such hierarchical GNNs are able to discover and adhere \cite{ying2018hierarchical} the latent hierarchical structure present in most of the real world networks \cite{newman2003structure}. Typically, these GNNs for graph classification feed the vector representation of the entire graph to a softmax layer and learn all the parameters of the network in an integrated and supervised way.

One limitation of most of the existing graph neural networks for graph level task is their dependency on the availability of graph labels. They use the labels to learn the parameters of GNN by minimizing a graph classification loss such as cross entropy of the predicted labels and the actual labels. Graphs having different labels are generally mapped to different regions in the embedding space.
A set of popular GNNs \cite{xu2018how,NIPS2019_Provably,ying2018hierarchical} use 80\% to 90\% graph labels in a dataset to train the respective neural architectures. But in real life situations, obtaining the ground truth labels of the graphs is expensive and often, not available. Thus, there is a strong need to design graph neural networks to generate unsupervised vector representation of an entire graph in a graph dataset, which can then be used for multiple graph downstream tasks.

There exist some unsupervised way of generating graph similarities and embedding. Graph kernel methods \cite{vishwanathan2010graph,shervashidze2011weisfeiler,yanardag2015deep}, which generate a kernel for a set of graphs in an unsupervised way, had been the state-of-the-art for graph classification for a long time. Typically, such a graph kernel is fed to a supervised classifier such as support vector machine for graph classification. Graph kernel methods use hand-crafted features and thus, often fail to adapt the data distribution present in the graph. There are also methods \cite{narayanan2017graph2vec,adhikari2018sub2vec} which use skip-gram \cite{mikolov2013distributed} type optimization to embed the entire graph to a vector space. But these methods are shallow in nature and do not exploit the power of graph neural network framework. Recently, unsupervised representation learning via maximizing mutual information between the output and different variants of input in the context of neural network achieves promising results for images \cite{belghazi2018mutual,hjelm2018learning}. The concept of information maximization is adopted to generate unsupervised node representations in DGI \cite{velivckovic2018deep}. Very recently, from the list of accepted papers in ICLR 2020, we come to know about another work InfoGraph\footnote{Our work was developed independently of InfoGraph. Though the aims of both the papers are similar, the methodology adopted are different. Please see Section \ref{sec:differen} to find the differences.} \cite{Sun2020InfoGraph} which uses information maximization to generate unsupervised and also semi-supervised representations for the entire graph.

Our goal in this work is to design GNN based architecture to generate unsupervised representation of an entire graph in a graph dataset. Following are the contributions we make:
\begin{itemize}
    \item We propose an unsupervised algorithm - \underline{Gra}ph representation by \underline{P}eriphery and \underline{H}ierarchical information \underline{max}imization, referred as \textbf{GraPHmax}, which combines hierarchical graph neural network and mutual information maximization into single framework.
    \item We propose periphery representation of a graph (different from the classical concept of graph periphery from graph theory \cite{west2001introduction}) and use it, along with the latent graph hierarchies to learn an unsupervised vector representation of a graph by maximizing two different mutual information - peripheral information and hierarchical information, in the proposed algorithm GraPHmax.
    \item We conduct thorough experiments on three downstream graph level tasks - graph classification, graph clustering and graph visualization to check the quality of the representations. We are able to improve the state-of-the-art with respect to both unsupervised and supervised algorithms on multiple real-world and popularly used graph datasets. Further, results on detailed model ablation study show the importance of individual components used in GraPHmax. 
\end{itemize}

\section{Related Work and Research Gaps}\label{sec:related}
A survey on network representation learning and graph neural networks can be found in \cite{wu2019comprehensive}. For the interest of space, we briefly discuss some more prominent approaches on representation learning for the entire graph. Graph kernel based approaches \cite{vishwanathan2010graph}, which map the graphs to Hilbert space implicitly or explicitly, remain to be the state-of-the-art for graph classification for long time. There are different types of graph kernels present in the literature, such as
random walk based kernel \cite{kashima2003marginalized}, 
Weisfeiler-Lehman subtree kernel \cite{shervashidze2011weisfeiler} and Deep graph kernel \cite{yanardag2015deep}. But most of the existing graph kernels use hand-crafted features and also ignore the node attributes present in a graph. So, they face difficulties to generalize well.

    Significant progress happened in the domain of node representation and node level tasks via graph neural networks. Spectral graph convolutional neural networks with fast localized convolutions \cite{defferrard2016convolutional,kipf2016semi},
position aware graph neural networks \cite{you2019position}
are some notable examples of GNN for node representation.
As mentioned in Section \ref{sec:intro}, different graph pooling strategies are present in the literature to go from node embeddings to a single representation for the whole graph \cite{duvenaud2015convolutional,gilmer2017neural,zhang2018end}.
DIFFPOOL \cite{ying2018hierarchical} is a recently proposed hierarchical GNN which uses a GCN based pooling to create a set of hierarchical graphs in each level. A self attention based hierarchical pooling strategy which determines the importance of a node to find the label of the graph is proposed in \cite{lee2019self}.
A theoretically most powerful (in terms of graph representation) first order GNN is proposed and referred as GIN in \cite{xu2018how}, which is further enhanced in \cite{NIPS2019_Provably}.
Higher order GNNs which operate beyond immediate neighborhood are also studied recently \cite{morris2019weisfeiler,abu2019mixhop}.

The above GNN methods that generate graph representation need large number of graph labels to learn the parameters of the neural networks through a graph classification loss. To overcome this, some unsupervised graph level embedding approaches are also proposed in the literature. The simplest approach can be to directly aggregate (for e.g., taking mean) the unsupervised node embeddings to obtain an unsupervised graph embedding. But this approach fails to discriminate graphs with different structures as unsupervised node embedding approaches do not consider the dynamics across different graphs \cite{grover2016node2vec}.  Inspired by the representation of documents in natural language processing literature \cite{le2014distributed}, graph2vec \cite{narayanan:mlg2017} is proposed which creates a vocabulary of subgraphs and then generates graph embeddings by maximizing the likelihood of the set of subgraphs which belong to a graph. Sub2Vec \cite{adhikari2018sub2vec} is another approach in the same direction which can generate embedding of arbitrary subgraphs and graphs by maximizing skip-gram type of objective. Recently, use of mutual information maximization in the context of unsupervised deep representation learning for images is studied in \cite{belghazi2018mutual,hjelm2018learning}. DGI \cite{velivckovic2018deep} is the first approach on graphs which maximize the mutual information between a graph level summary and individual nodes to obtain node representations. Very recently, InfoGraph \cite{Sun2020InfoGraph} is proposed to map an entire graph to a vector space by maximizing mutual information through graph neural network. To the best of our knowledge, InfoGraph is the first GNN based unsupervised algorithm to learn graph level representation. But, both DGI and InfoGraph do not adhere the hierarchical structure present in most of the real world graphs. Our work precisely addresses this gap in the literature and enhances the discriminating power of graph representation by maximizing information through different hierarchies. Section \ref{sec:differen} presents a detailed differentiation of our proposed algorithm with some close related work in this direction.

\section{Problem Formulation}\label{sec:prob}
In unsupervised graph representation, a set of graphs are given and the objective is to map each graph to a vector, so that similar graphs are close in the vector space and dissimilar graphs are far apart. These types of vector representation of a graph can be used in multiple downstream graph level machine learning tasks. 
More formally, given a set of $M$ graphs $\mathcal{G} = \{G_1, G_2, \cdots, G_M\}$, our goal is to learn a function $f: \mathcal{G} \mapsto \mathbb{R}^K$ which maps each graph $G \in \mathcal{G}$  to a $K$-dimensional vector representation $x^{G}$.

Each graph $G_i$ can be defined as $G_i = (V_i,E_i,X_i)$ where $V_i$ is the set of $n_i$ nodes and $E_i$ is the set of edges. Each node $v^i_j$ is associated with a attribute vector $x^i_j \in \mathbb{R}^D$. Please note, we have used $x$ to denote both the final graph level representation and the node features. But we eliminate the ambiguity by using different subscripts and superscripts appropriately. We use both the link structure and the node attributes of the graphs to obtain their final representations in an unsupervised way.

\begin{figure}[h!]
  \centering
  \begin{subfigure}[b]{0.3\linewidth}
    \includegraphics[width=\linewidth]{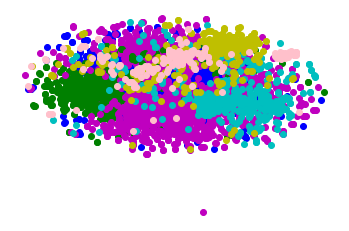}
    \caption{}
  \end{subfigure}
  \begin{subfigure}[b]{0.3\linewidth}
    \includegraphics[width=\linewidth]{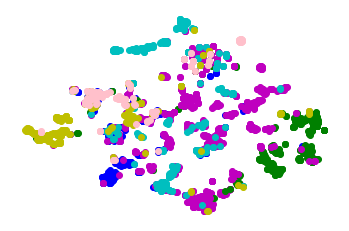}
    \caption{}
  \end{subfigure}
  \begin{subfigure}[b]{0.3\linewidth}
    \includegraphics[width=\linewidth]{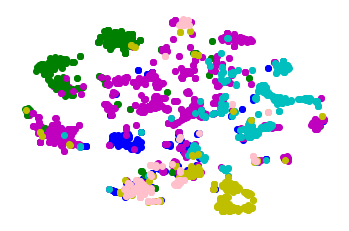}
    \caption{}
  \end{subfigure}
  \caption{t-SNE \cite{vanDerMaaten2008} visualization of the nodes of CORA using: (a) initial node features i.e. X; (b) node embeddings from GIN encoder i.e. H; (c) Periphery node representations i.e., X-H. Different colors represent different node labels.}
  \label{fig:nodeViz}

\end{figure}

\section{Solution Approach}\label{sec:soln}
\begin{figure*}[h!]
  \centering
    \includegraphics[width=\linewidth]{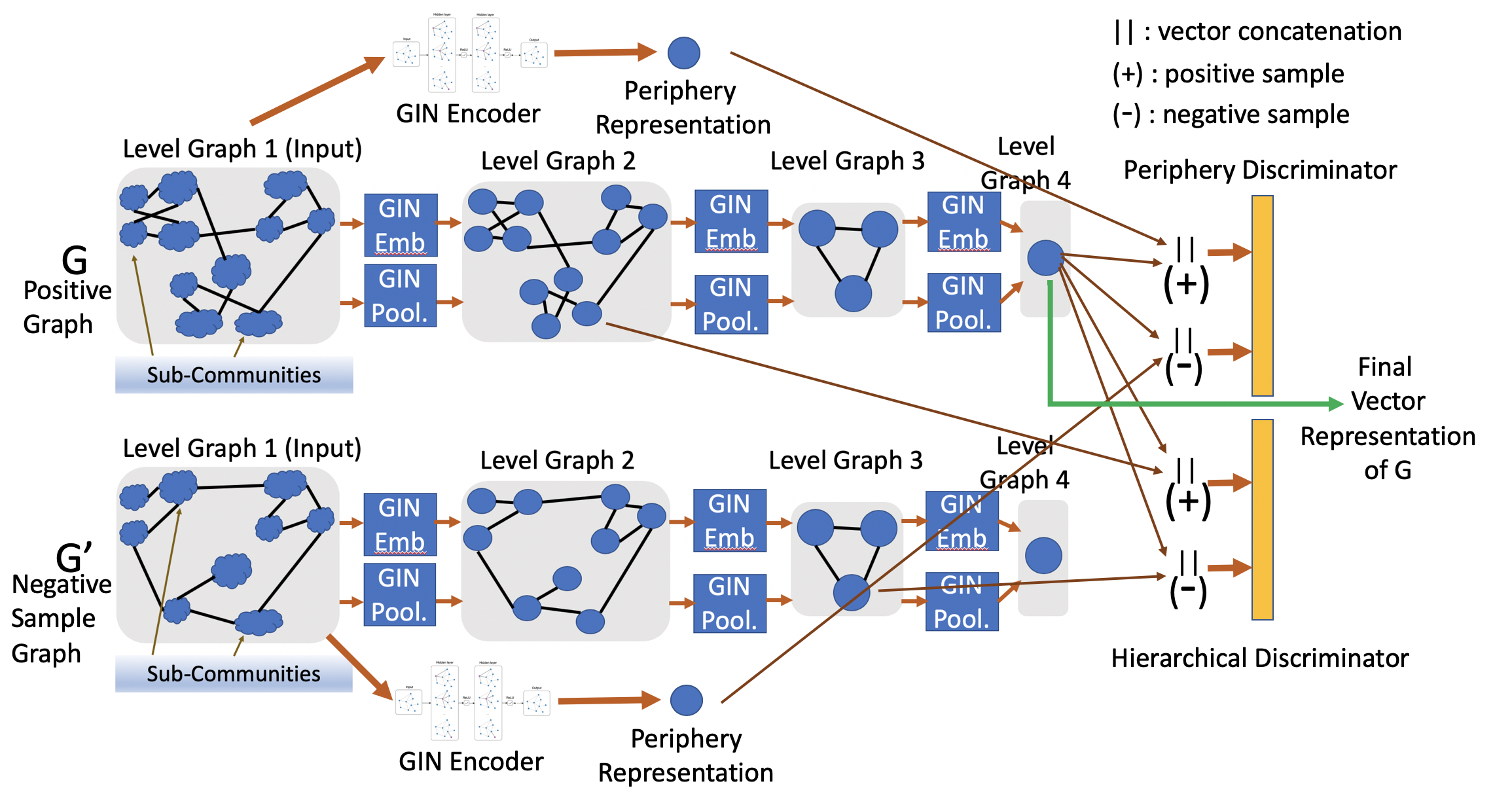}
    \caption{Architecture of GraPHmax for unsupervised graph representation}
    \label{fig:GraPHmax}
\end{figure*}

There are multiple components of the proposed solution GraPHmax, as shown in Figure \ref{fig:GraPHmax}. We explain each of them below.

\subsection{Periphery Representation}\label{sec:periphery}
As mentioned in Section \ref{sec:intro}, different graphs in a graph dataset inherently borrow some common functionalities. For example, protein-protein interaction networks \cite{krogan2006global} are mathematical representations of the physical contacts between proteins in the cell. Naturally, each graph in the dataset would certainly contain the fundamental behavior of protein interaction irrespective of their labels. Similarly, in any social network \cite{scott1988social}, some core properties like small world effect (e.g., "six degrees of separation"), scale-freeness can be observed irrespective of its label. Graph compression on the adjacency matrix (by matrix factorization or by autoencoders) often captures these core properties of the network \cite{wang2012nonnegative}, which are required to reconstruct the network. Such compressed representations are used for node embedding in multiple works \cite{wang2016structural,bandyopadhyay2019outlier}. But for graph representation, one may need more discriminative features, for which we introduce the concept of the periphery representation of a graph. Please note, the concept of periphery representation of a graph introduced here is technically different from the classical concept of graph periphery, which is the set of nodes of a graph with eccentricity equal to the diameter of the graph \cite{west2001introduction}. However, there is conceptual similarity between them as we consider the outskirt representation of a graph as the periphery representation which is discussed below.

Here we describe the methodology to obtain the periphery representation of the graph. Given a graph $G=(V,E,X)$, we use a GIN \cite{xu2018how} encoder whose $l$th layer can be defined as:.
\begin{equation}\label{eq:GIN}
    h_v^{l+1} = MLP^l \Big( (1+\epsilon^l) h_v^l + \sum\limits_{u \in \mathcal{N}(v)} h_u^l \Big) 
\end{equation}
Here, $h_u^l$ is the representation of the node $u$ in $l$th GIN layer.
To compute $h_v^{l+1}$, i.e., the representation of the node $v$ in $l+1$th layer of GIN, it takes the sum of the features from all the neighbor nodes and then take a weighted sum of that with its own features. $\epsilon$ is a learnable parameter. MLP is a multi-layer perceptron with non-linear activation function. Hidden representation $h_v^0$ before the first GIN layer for any node $v$ is initialized with the node attributes from the matrix $X$. GIN has been shown to be the most powerful 1st order GNN for graph representation \cite{xu2018how}.

We keep the dimension of the final node embedding matrix $H$ from the GIN encoder to be the same as the input feature dimension, i.e., $H \in \mathbb{R}^{|V| \times D}$. Then we use a cross entropy loss between the adjacency matrix $A$ of the graph and $HH^T$. Clearly, $HH^T$ captures the node similarity through the hidden representations of the nodes and we want that to be coherent with the link structure of the graph which is captured through the adjacency matrix. Thus, once the training is complete, $H$ captures the features of a node along with its higher order neighbors (depending on the number of layers used in the GIN encoder) to reconstruct the adjacency structure of the graph. As explained before, we want to capture the periphery representation of the nodes, rather than its core representation. So, we define the periphery representation matrix $S \in \mathbb{R}^{|V| \times D}$ as:
\begin{equation}
    S = X - H
\end{equation}
The intuition behind this definition is as follows. Initial features of a node, along with the link structure contains two parts. First, the core representation which is essential to reconstruct the network. Second, the periphery representation which somewhat captures the outskirt of the graph and less important for graph reconstruction.
So in the vector space, we subtract the core node embedding features (EF, i.e., $H$) from the initial node features (NF, i.e., $X$) to obtain the periphery representations. Experimentally, it turns out that such a simple definition of periphery representation is quite powerful to use in the later modules for obtaining an entire graph level representation, as shown in Section \ref{sec:exp}.
Further, the discrimination power of periphery node representation over NF and EF are shown in Figure \ref{fig:nodeViz} on the dataset Cora \cite{kipf2016semi} which is a citation network.

\begin{table}
\centering
\begin{tabular}{*6c}
	\toprule
	\sffamily{Notations} & Explanations\\
    \hline
	\midrule
    $\mathcal{G}=\{G_1,\cdots,G_M\}$ & Set of graphs in a graph dataset \\x` 
    $G=(V,E,X)$ & One single graph \\
    $x^i_j \in \mathbb{R}^D$ & Attribute vector for $j$th node in $i$th graph \\
    $x^G \in \mathbb{R}^K$ & Final representation of the graph $G$ \\
    $S \in \mathbb{R}^{|V| \times D}$ & Periphery nodes representations of $G$ \\
    $G^1,\cdots,G^R$ & Level graphs of some input graph $G$ \\
    $Z_r \in \mathbb{R}^{N_r \times K}$ & Embedding matrix of $G^r$ \\
    $P_r \in \mathbb{R}^{N_r \times N_{r+1}}$ & Node assignment matrix from $G^r$ to $G^{r+1}$ \\
    $\mathcal{D}^P$, $\mathcal{D}^H$ & Periphery and hierarchical discriminators \\
    \bottomrule
	\end{tabular}
\caption{Different notations used in the paper}
\label{tab:datasets}
\end{table}

\subsection{Hierarchical Graph Representation}\label{sec:hierar}
Graphs exhibit hierarchical structure by nature \cite{newman2003structure}. For example, an entire graph can be divided into multiple communities, each community may have sub-communities within it. A sub-community can be formed by multiple nodes. Communities (and also sub-communities) are inter-lined between them and can also be overlapping \cite{gregory2010finding}. Graph representation mechanisms should discover and adhere such latent hierarchies in the graph. Recently, hierarchical graph neural networks have been shown to be promising for supervised graph classification \cite{ying2018hierarchical,lee2019self}. The idea of hierarchical GNN is extended in this work for unsupervised graph representation, by integrating it with the information maximization mechanisms.
This subsection discusses the hierarchical structure of GraPHmax. 

As shown in Figure \ref{fig:GraPHmax}, there are $R=4$ different levels of the graph in the hierarchical architecture. The first level is the input graph. Let us denote these level graphs (i.e., graphs at different levels) by $G^1, \cdots, G^R$. There is a GNN layer between the level graph $G^r$ (i.e., the graph at level $r$) and the level graph $G^{r+1}$. This GNN layer comprises of an embedding layer which generates the embedding of the nodes of $G^r$ and a pooling layer which maps the nodes of $G^r$ to the nodes of $G^{r+1}$. We refer the GNN layer between the level graph $G^r$ and $G^{r+1}$ by $r$th layer of GNN, $\forall r=1,2,\cdots,R-1$.
Please note, number of nodes $N_1$ in the first level graph depends on the input graph, but we keep the number of nodes $N_r$ in the consequent level graphs $G^r$ ($\forall r=2,\cdots,R$) fixed for all the input graphs (in a dataset of graphs).
As pooling mechanisms shrink a graph, $N_r>N_{r+1}$, $\forall r \leq R-1$. To obtain a single vector representation for the entire graph, we keep only one node in the last level graph, i.e., $N_{R}=1$.

Let us assume that any level graph $G^r$ is defined by its adjacency matrix $A_r \in \mathbb{R}^{N_r \times N_r}$ and the feature matrix $X_r \in \mathbb{R}^{N_r \times K}$ (except for $G^1$, which is the input graph and its feature matrix $X_1 \in \mathbb{R}^{N_1 \times D}$). 
The $r$th embedding layer and the pooling layer are defined by:
\begin{flalign}\label{eq:embPool}
    Z_r = \text{GIN}_{r,embed}(A_r,X_r) \nonumber \\
    P_r = \text{softmax}(\text{GIN}_{r,pool}(A_r,X_r))
\end{flalign}
Here, $Z_r \in \mathbb{R}^{N_r \times K}$ is the embedding matrix of the nodes of $G^r$. The softmax after the pooling is applied row-wise. $(i,j)$th element of $P_r \in \mathbb{R}^{N_r \times N_{r+1}}$ gives the probability of assigning node $v_i^r$ in $G^r$ to node $v_j^{r+1}$ in $G^{r+1}$. Based on these, the graph $G^{r+1}$ is constructed as follows,
\begin{flalign}\label{eq:adjFeat}
    A_{r+1} = P_r^T A_r P_r \in \mathbb{R}^{N_{r+1} \times N_{r+1}} 
    \;;\;\nonumber \\ 
    X_{r+1} = P_r^T Z_r \in \mathbb{R}^{N_{r+1} \times K}
\end{flalign}
The matrix $P_r$ contains information about how nodes in $G^r$ are mapped to the nodes of $G^{r+1}$, and the adjacency matrix $A_r$ contains information about the connection of nodes in $G^r$. Eq. \ref{eq:adjFeat} combines them to generate the connections between the nodes (i.e., the adjacency matrix $A_{r+1}$) of $G^{r+1}$. Node feature matrix $X_{r+1}$ of $G^{r+1}$ is also generated similarly.
As the embedding and pooling GNNs, we use GIN \cite{xu2018how} as it is shown to be the most powerful 1st order GNN. The update rule of GIN is already presented in Equation \ref{eq:GIN}. $X_R$ gives the representation of the single node in the last level graph, which we consider as the representation of the entire graph and train it in the information maximization mechanism discussed next.

\subsection{Information Maximization}\label{sec:infomax}
In this section, we aim to obtain a graph level representation by maximizing the mutual information of the graph level representation to the periphery representation and the representations in different levels of the hierarchy. 
%
We adopt the concept of training a classifier to distinguish between samples coming from the joint distribution and the product of marginals \cite{belghazi2018mutual}. First, let us use the notation $x^G = X_R^T \in \mathbb{R}^K$ to denote the graph level representation obtained from the hierarchical module of GraPHmax. We maximize the mutual information between this graph level representation with respect to (i) periphery representation obtained in Section \ref{sec:periphery}, and (ii) different node representations in the intermediate level graphs of the hierarchical representation. We explain them below.

\subsubsection{Periphery Information Maximization}\label{sec:pinfomax}
As explained in Section \ref{sec:periphery}, we obtain the periphery node representation matrix $S$ for a graph $G$. To obtain a graph level representation, we take the mean of the node representations (i.e., the mean of the rows of $S$) and denote it as $s^G$. To maximize the mutual information, we maximize the following noise-contrasive objective with a binary cross entropy loss between the samples from the joint (positive examples) and the product of the marginals (negative examples) \cite{hjelm2018learning,velivckovic2018deep}.
\begin{equation}\label{eq:infoPeri}
    \mathcal{L}^P = \sum\limits_{G \in \mathcal{G}} \bigg[ log \; \mathcal{D}^P \Big( x^G, s^G \Big) + \mathbb{E}_{G' \thicksim \bar{\mathbb{P}}(\mathcal{G})} \; \Big[ log \; \Big( 1 - \mathcal{D}^P \Big( x^G, s^{G'} \Big) \Big) \Big] \bigg]
\end{equation}
Here $\mathcal{D}^P$ is a discriminator, the subscript $P$ stands for periphery. We use a logistic regression classifier here. For each graph $G$ in the set of graphs $\mathcal{G}$, $\mathcal{D}^P$ tries to maximize the probability when the entire graph representation $x^G$ and the periphery graph representation $s^G$ come from the same graph $G$. 
Here, $\bar{\mathbb{P}}(\mathcal{G})$ is the negative sampling distribution on the set of graphs of the noise-contrasive objective.
In our implementation, for each selected positive graph sample, $n_P > 0$ (number of negative samples for periphery discriminator) negative graphs are sampled randomly from $\mathcal{G}$ to evaluate the expectation term $\mathbb{E}_{G' \thicksim \bar{\mathbb{P}}(\mathcal{G})} \; \Big[ log \; \Big( 1 - \mathcal{D}^P \Big( x^G, s^{G'} \Big) \Big) \Big] \bigg]$ in Eq. \ref{eq:infoPeri}.
For each such negative graph $G'$, $\mathcal{D}^P$ tries to minimize the probability that $x^G$ and $s^{G'}$ comes from the same graph.

Maximizing the above objective function would ensure the final graph representation, which is an output of the hierarchical GNN structure, preserves high mutual information with the periphery representation, which is an outskirt representation of the same graph. The use of noise-contrasive objective also ensures that representation of a graph should have less mutual information with the periphery representation of the other graphs in the dataset. As periphery representations are quite discriminative in nature, maximizing the above objective leads to obtaining a discriminative unsupervised graph representations for different graphs.

\subsubsection{Hierarchical Information Maximization}\label{sec:hinfomax}
Periphery information maximization helps to exploit the complementary information available in the periphery and the hierarchical representation of a graph. But, as shown in \cite{hjelm2018learning}, maximizing information between global features and local features often helps to improve the quality of unsupervised representation. But unlike images, input graphs in a graph dataset can vary in size and thus preserving mutual information between the entire graph level representation and to that of all the individual nodes is computationally expensive. Beside, the role of a graph is often determined by the higher level structures and their inter connections (such as arrangement of communities in a graph, presence of a subgraph etc. \cite{morris2019weisfeiler}). As we already employ a hierarchical GNN structure in GraPHmax, instead of individual nodes of a graph, we decide to preserve the mutual information of the entire graph level representation with the representations of the nodes within different intermediate level graphs (i.e., with $2 \leq r \leq R-1$) of GraPHmax.

Let us introduce the notation $\mathcal{H}(G)= \{v \in V(G^r) \; | \; 2 \leq r \leq R-1 \}$ to denote the set of nodes in intermediate levels graphs in the hierarchical representation of the graph $G$. Similar to the case of periphery information maximization, we again maximize the following noise-contrasive objective with a binary cross entropy loss to maximize the mutual information between the entire graph level representation to that of the nodes in the intermediate level graphs of the hierarchical graph representation.
\begin{align}\label{eq:infoHier}
    \mathcal{L}^H = & \sum\limits_{G \in \mathcal{G}} \bigg[ \sum\limits_{v \in \mathcal{H}(G)} log \; \mathcal{D}^H \Big( x^G, Z^G_v \Big) \\ \nonumber
    & + \mathbb{E}_{G' \thicksim \bar{\mathbb{P}}(\mathcal{G})} \; \Big[ \sum\limits_{v \in \mathcal{H}(G')} log \; \Big( 1 - \mathcal{D}^H \Big( x^G, Z^{G'}_v \Big) \Big) \Big] \bigg]
\end{align}
Here, $Z^G_v$ is the embedding of the node $v$ from some intermediate level graph of the hierarchical representation of $G$, as shown in Section \ref{sec:hierar}. $Z^G_v$ can be obtained from Eq. \ref{eq:embPool}. $\mathcal{D}^H$ is the logistic regression based classifier which assigns high probability when $x^G$ and $Z^G_v$ comes from the same graph. Similar to the case of periphery information maximization, for each selected positive graph sample, $n_H > 0$ (number of negative samples for hierarchical discriminator) negative graphs are sampled randomly from $\mathcal{G}$ to evaluate the expectation term in Eq. \ref{eq:infoHier}. The discriminator $\mathcal{D}^P$ tries to minimize the probability of the pair $x^G$ and $Z^{G'}_v$ for such a negative graph $G'$. Please note, for any graph $G$, we have considered all the intermediate nodes ($\in \mathcal{H}(G)$) from the hierarchical representation of $G$. For larger graphs, one can also sample to save computation time. A detailed study on the effect of such a sampling strategy can be conducted in future. By preserving the mutual information between different intermediate nodes and the final graph embedding of the hierarchical graph representation, the entire graph representation would be able to contain the similarities between different hierarchies of the graph\footnote{Similar observation was made in \cite{hjelm2018learning} for patches in image representation}.

\subsection{Analysis of GraPHmax}\label{sec:keyAna}
\subsubsection{Training of GraPHmax}\label{sec:training}
First, we train the GIN encoder to generate the periphery representation by minimizing the cross entropy loss of graph reconstruction. Once the periphery representations are obtained, we start training the parameters of the hierarchical graph representation unit and the two discriminators by maximizing the objectives in Equations \ref{eq:infoPeri} and \ref{eq:infoHier}. We use ADAM \cite{DBLP:journals/corr/KingmaB14} optimization in the backpropagation algorithm. Details of parameterization are explained in Section \ref{sec:expSetup} and Appendix \ref{sec:supp}.

\subsubsection{Model Complexity}\label{sec:complexity}
GraPHmax consists of multiple modules. The GIN encoder to generate periphery representation has $O(D^2)$ number of parameters to learn. In the hierarchical architecture, which have multiple GIN embedding and pooling layers, have a total of $O(DK + (R-2)K^2)$ parameters. This is because only the first level graph (i.e., the input graph) has $D$ dimensional feature vector in each node, and all other graphs have $K$ dimensional feature vectors. Finally, the discriminators have another $O(D+K)$ parameters to train. Hence, the total number of parameters to train in GraPHmax is independent of the number of graphs in the dataset, and also the number of nodes present into those graphs. The learning in GraPHmax is stable because of having less number of parameters to train, which we observe in the experiments as well. 

\subsubsection{Differentiating with some recently proposed GNNs}\label{sec:differen}
We differentiate GraPHmax from the three key related works.

\textbf{DIFFPOOL} \cite{ying2018hierarchical}: DIFFPOOL proposes a GNN based supervised hierarchical graph classification architecture. GraPHmax also has a hierarchical graph representation module present in in. But there are certain differences between the two. First, in terms of design, DIFFPOOL uses GCN \cite{kipf2016semi} embedding and pooling layers. Whereas, the hierarchical module of GraPHmax employs GIN \cite{xu2018how} as embedding and pooling layers, because of the theoretical guarantee on the representative power of GIN. Second, DIFFPOOL is a supervised (graph classification) technique. Whereas, GraPHmax is completely unsupervised. Naturally, there are multiple other modules in GraPHmax to make the unsupervised learning efficient.

\textbf{DGI} \cite{velivckovic2018deep}: DGI maximizes the mutual information between a graph level summary and the nodes of the graph to learn the node representations. Whereas, GraPHmax generates unsupervised embedding of an entire graph in a set of graphs. The methodology adopted (periphery and hierarchical information maximization) in GraPHmax is different from DGI.

\textbf{InfoGraph} \cite{Sun2020InfoGraph}: InfoGraph maximizes mutual information between a graph level representation and its nodes. Authors concatenate the embedding of different layers of a GNN to capture higher order dependency (i.e., beyond immediate neighborhood) among the nodes. But communities in a graph can be of varying size and extent. Covering same depth through GNN from every node would not be able to adhere the community (or sub-community) structure of the graph. Whereas, the use of hierarchical GNN in GraPHmax explicitly discovers the latent hierarchical structure in a graph. We also maximize the mutual information between periphery and different units of hierarchical units of the graph to that of the entire graph representation, which is not the case in InfoGraph.

Further, we compare the performance of GraPHmax with all these algorithms in Section \ref{sec:exp} on real world graph datasets.

\begin{table}
\centering
\begin{tabular}{*6c}
	\toprule
	\sffamily{Dataset} & \#Graphs & \#Max Nodes & \#Labels & \#Attributes\\
    \hline
	\midrule
	\sffamily{MUTAG} & 188 & 28 & 2 & NA\\
	\sffamily{PTC} & 344 & 64 & 2 & NA\\
    \sffamily{PROTEINS} & 1113 & 620 & 2 & 29\\
    \sffamily{NCI1} & 4110 & 111 & 2 & NA\\
    \sffamily{NCI109} & 4127 & 111 & 2 & NA\\
    \sffamily{IMDB-BINARY} & 1000 & 136 & 2 & NA\\
    \sffamily{IMDB-MULTI} & 1500 & 89 & 3 & NA\\
    
    \bottomrule
	\end{tabular}
\caption{Different datasets used in our experiments}
\label{tab:datasets}
\end{table}

\subsection{Model Ablation Study of GraPHmax}\label{sec:ablation}
There are multiple modules present in the overall architecture of GraPHmax. Thus we present the following variants of GraPHmax to conduct a thorough model ablation study.

\textbf{GraPHmax+NF}: Here we replace the periphery representation of the nodes with just the initial node features (NF) from Section \ref{sec:periphery}. Hence, the graph representation along with the mean node features are given as the input to the periphery discriminator. Comparison to GraPHmax+RF shows the usefulness of periphery representation of the graph for obtaining a final graph level representation.

\textbf{GraPHmax+EF}: Here we replace the periphery representations of the nodes with the node embedding matrix $H$ (EF stands for embedding features) as obtained directly from Equation \ref{eq:GIN}, while all other modules of GraPHmax remain intact. This also shows the usefulness of periphery representation in GraPHmax.

\textbf{GraPHmax-P}: Here we remove the periphery representation learning module and the periphery discriminator from the overall architecture of GraPHmax. Hence, a graph representation is obtained through the hierarchical architecture and hierarchical mutual information maximization.

\textbf{GraPHmax-H}: Here we remove the hierarchical discriminator from GraPHmax. Hence, a graph representation is learned through hierarchical architecture and periphery information maximization.

We compare the performance of GraPHmax with each of these variants to show the usefulness of respective modules of GraPHmax.

\begin{table*}
\centering
\begin{tabular}{*8c} 
	\toprule
	\sffamily{Algorithms} & \footnotesize MUTAG&PTC&PROTEINS&NCI1&NCI109&IMDB-B&IMDB-M \\
    \hline
	\midrule
    
    GK \cite{shervashidze2009efficient}  &81.39$\pm$1.7&55.65$\pm$0.5&71.39$\pm$0.3&62.49$\pm$0.3&62.35$\pm$0.3&NA&NA \\
    RW \cite{vishwanathan2010graph} &79.17$\pm$2.1&55.91$\pm$0.3&59.57$\pm$0.1&NA& NA&NA&NA\\
    PK \cite{neumann2016propagation} &76$\pm$2.7&59.5$\pm$2.4&73.68$\pm$0.7&82.54$\pm$0.5&NA&NA&NA\\
    WL \cite{shervashidze2011weisfeiler}  &84.11$\pm$1.9&57.97$\pm$2.5&74.68$\pm$0.5&\textbf{84.46$\pm$0.5}&\textbf{85.12$\pm$0.3}&NA&NA\\
    AWE-DD \cite{ivanov2018anonymous} &NA& NA& NA& NA& NA&74.45$\pm$5.8&51.54$\pm$3.6\\
    AWE-FB \cite{ivanov2018anonymous} &87.87$\pm$9.7& NA &NA& NA& NA&73.13$\pm$3.2&51.58$\pm$4.6\\
    \hline
    
    DGCNN \cite{zhang2018end} & 85.83$\pm$1.7& 58.59$\pm$2.5& 75.54$\pm$0.9& 74.44$\pm$0.5& NA&70.03$\pm$0.9&47.83$\pm$0.9\\
    PSCN \cite{niepert2016learning} &88.95$\pm$4.4&62.29$\pm$5.7&75$\pm$2.5&76.34$\pm$1.7&NA&71$\pm$2.3&45.23$\pm$2.8\\
    DCNN \cite{atwood2016diffusion} &NA&NA&61.29$\pm$1.6&56.61$\pm$1.0&NA&49.06$\pm$1.4&33.49$\pm$1.4\\
    ECC \cite{simonovsky2017dynamic} &76.11&NA&NA&76.82&75.03&NA&NA\\
    DGK \cite{yanardag2015deep} &87.44$\pm$2.7&60.08$\pm$2.6&75.68$\pm$0.5&80.31$\pm$0.5&80.32$\pm$0.3&66.96$\pm$0.6&44.55$\pm$0.5\\
    DiffPool \cite{ying2018hierarchical} &79.50$\pm$5.8&NA&76.25&NA&NA&NA&NA\\
    IGN \cite{maron2018invariant} &83.89$\pm$12.95&58.53$\pm$6.86&76.58$\pm$5.49&74.33$\pm$2.71&72.82$\pm$1.45&72.0$\pm$5.54&48.73$\pm$3.41\\
    GIN \cite{xu2018how} &89.4$\pm$5.6&64.6$\pm$7.0&76.2$\pm$2.8&82.7$\pm$1.7&NA&\textbf{75.1$\pm$5.1}&\textbf{52.3$\pm$2.8}\\
    1-2-3GNN \cite{morris2019weisfeiler} &86.1$\pm$&60.9$\pm$&75.5$\pm$&76.2$\pm$&NA&74.2$\pm$&49.5$\pm$\\
    3WL-GNN \cite{maron2019provably} &90.55$\pm$8.7&66.17$\pm$6.54&\textbf{77.2$\pm$4.73}&83.19$\pm$1.11&81.84$\pm$1.85&72.6$\pm$4.9&50$\pm$3.15\\
    \hline
    
    node2vec \cite{grover2016node2vec} &72.63$\pm$10.20&58.85$\pm$8.00&57.49$\pm$3.57&54.89$\pm$1.61&52.68$\pm$1.56&NA&NA\\
    sub2vec \cite{narayanan2016subgraph2vec} &61.05$\pm$15.79&59.99$\pm$6.38&53.03$\pm$5.55&52.84$\pm$1.47&50.67$\pm$1.50&55.26$\pm$1.54&36.67$\pm$0.83\\
    graph2vec \cite{narayanan2017graph2vec} &83.15$\pm$9.25&60.17$\pm$6.86&73.30$\pm$2.05&73.22$\pm$1.81&74.26$\pm$1.47&71.1$\pm$0.54&50.44$\pm$0.87\\
    DGI \cite{velivckovic2018deep}&89.36$\pm$4.16&60.78$\pm$9.80&73.23$\pm$4.38&67.85$\pm$1.97&68.69$\pm$0.86&73.99$\pm$4.97&50.66$\pm$3.46\\
    InfoGraph \cite{Sun2020InfoGraph} &89.01$\pm$1.13&61.65$\pm$1.43&NA&74.20$\pm$2.11&72.59$\pm$1.96&73.03$\pm$0.87&49.69$\pm$0.53\\
    \hline
    GraPHmax+NF&89.47$\pm$8.30&63.12$\pm$6.25&73.12$\pm$3.61&73.76$\pm$3.15&72.64$\pm$1.36&71.99$\pm$6.35&51.93$\pm$3.75\\
    GraPHmax+EF&92.44$\pm$5.08&63.43$\pm$5.91&73.91$\pm$2.03&74.19$\pm$2.68&73.18$\pm$1.58&72.30$\pm$6.99 &50.73$\pm$4.02\\
    GraPHmax-P&89.33$\pm$5.19&60.60$\pm$8.10&74.39$\pm$4.94&72.74$\pm$2.22&72.74$\pm$2.22&69.80$\pm$4.81&49.40$\pm$3.58\\
    GraPHmax-H&90.50$\pm$7.65&57.28$\pm$3.32&64.60$\pm$3.73&72.5$\pm$3.95&72.96$\pm$2.68&71.90$\pm$7.11&45.99$\pm$3.55\\
    \textbf{GraPHmax}&\textbf{94.22}$\pm$4.33&\textbf{68.02$\pm$4.49}&76.11$\pm$1.81&75.51$\pm$3.34&74.94$\pm$2.06&74.69$\pm$6.13&51.93$\pm$3.41\\
    Rank &1&1&5&9&5&2&2 \\
    \bottomrule
\end{tabular}
\caption{Classification accuracy (\%) of different algorithms (21 state-of-the-art baselines and the proposed algorithm GraPHmax along with its 4 variants) for graph classification. NA denotes the case when the result of a baseline algorithm could not be found on that particular dataset from the existing literature. The last row `Rank' is the overall rank (1 being the highest position) of our proposed algorithm among all the algorithms present in the table. Please note, GraPHmax is able to consistently achieve the best performance in the unsupervised graph embedding category (last two classes of algorithms in the table).}
\label{tab:graphClassi}
\end{table*}

\section{Experimental Evaluation}\label{sec:exp}
We conduct thorough experiments in this section. Graph classification being the most important downstream graph level task, we present a detailed experimental setup before presenting the results on it. As our algorithm is unsupervised in nature, we do conduct experiment also on graph clustering and graph visualization.

\begin{table}
\centering
\begin{tabular}{*4c} 
	\toprule
	\sffamily{Algorithms} & \footnotesize MUTAG&PROTEINS&IMDB-M \\
    \hline
	\midrule
    DGI&72.34&\textbf{59.20}&33.66\\
    InfoGraph &77.65&58.93&33.93\\
    \hline
    GraPHmax+NF&84.10&57.14&36.40\\
    GraPHmax+EF&68.08&50.65&33.66\\
    GraPHmax-P&77.12&55.71&34.93\\
    GraPHmax-H&76.59&54.89&34.53\\
    \textbf{GraPHmax}&\textbf{85.04}&56.33&\textbf{37.00}\\
    \bottomrule
\end{tabular}
\caption{Clustering accuracy(\%) of unsupervised GNN based graph representation algorithms.}
\label{tab:graphClus}
\end{table}

\subsection{Setup for Graph Classification}\label{sec:expSetup}
We use a total of 7 graph datasets including 5 bioinformatics datasets (MUTAG, PTC, PROTEINS, NCI1 and NCI09) and 2 social network datasets (IMDB-BINARY and IMDB-MULTI). The details of these datasets can be found at (\url{https://bit.ly/39T079X}). Table \ref{tab:datasets} contains a high-level summary of these datasets.

We use a diverse set of state-of-the-art baselines. To evaluate the performance for graph classification, we compare GraPHMax (and its variants) with 21 baselines: 6 graph kernels, 10 supervised GNNs and 5 unsupervised graph representation algorithms (Table \ref{tab:graphClassi}). For node2vec \cite{grover2016node2vec}, we take the average of all the node embeddings as the graph embedding. For DGI \cite{velivckovic2018deep}, the graph level summary learned in the algorithm is considered as the graph representation. Along with that, we also consider 4 simpler variants of GraPHmax which are explained in model ablation study in Section \ref{sec:ablation}.

Periphery (GIN encoder) and Hierarchical representation modules of GraPHmax have different set of hyper-parameters and we explored various ranges to obtain the optimal value of these hyper-parameters. Number of layers (Intermediate layers and output layer) and number of units in each GIN layer of the GIN encoder were searched in \{2,3\} and \{64,128\} respectively for all the datasets. Please note that the dimension of last GIN layer (embedding layer) is same as the input dimension. There are few more hyperparameters for hierarchical representation part, the details of which are present in Section \ref{sec:sensitivity} and Appendix \ref{sec:supp}.
%
Additionally, we feed the unsupervised graph representations to a logistic regression (LR) classifier. The learning rate of LR is varied in \{0.1,0.01,0.05,0.001\} and fixed using 10-fold cross validation for all datasets, following the set-up as used in GIN \cite{xu2018how}. We report the averaged accuracy and corresponding standard deviation in table \ref{tab:graphClassi} . All the experiments are conducted on a shared server having Intel(R) Xeon(R) Gold 6142 processors running on CentOS7.6-Linux x86\_64Platform. 

\begin{figure*}[h!]
  \centering
  \begin{subfigure}[b]{0.19\linewidth}
    \includegraphics[width=\linewidth]{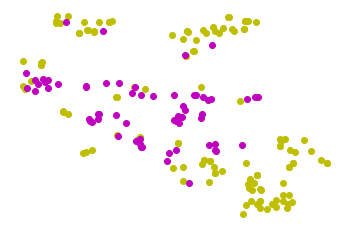}
    \caption{}
    \label{fig:EmbDim}
  \end{subfigure}
  \begin{subfigure}[b]{0.19\linewidth}
    \includegraphics[width=\linewidth]{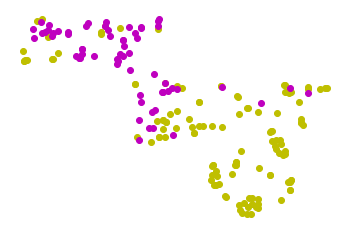}
    \caption{}
    \label{fig:EmbDim}
  \end{subfigure}
  \begin{subfigure}[b]{0.19\linewidth}
    \includegraphics[width=\linewidth]{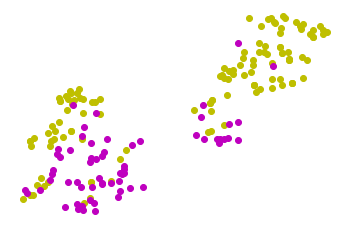}
    \caption{}
    \label{fig:EmbDim}
  \end{subfigure}
  \begin{subfigure}[b]{0.19\linewidth}
    \includegraphics[width=\linewidth]{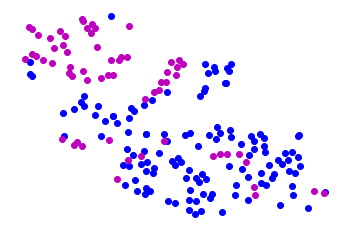}
    \caption{}
    \label{fig:EmbDim}
  \end{subfigure}
  \begin{subfigure}[b]{0.19\linewidth}
    \includegraphics[width=\linewidth]{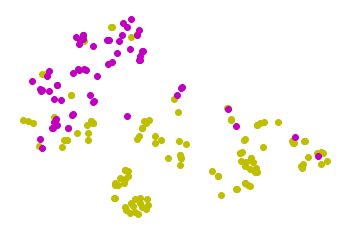}
    \caption{}
    \label{fig:EmbDim}
  \end{subfigure}
  \caption{t-SNE visualization of the graphs from MUTAG (different colors show different labels of the graphs) by the representations generated by: (a) GraPHmax+NF; (b) GraPHmax+EF; (c) GraPHmax-P (d) GraPHmax-H; and (e) GraPHmax.}
  \label{fig:sensityPTC}

\end{figure*}

\begin{figure*}[h!]
  \centering
  \begin{subfigure}[b]{0.19\linewidth}
    \includegraphics[width=\linewidth]{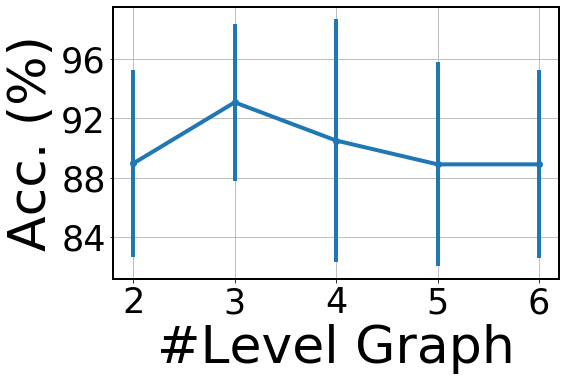}
    \caption{}
    \label{fig:subgraphSize}
  \end{subfigure}
  \begin{subfigure}[b]{0.19\linewidth}
    \includegraphics[width=\linewidth]{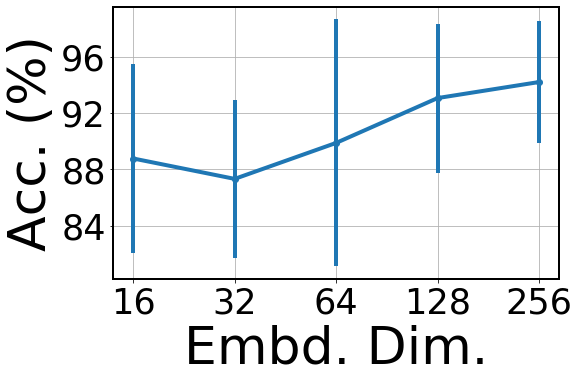}
    \caption{}
    \label{fig:NumSampled}
  \end{subfigure}
  \begin{subfigure}[b]{0.19\linewidth}
    \includegraphics[width=\linewidth]{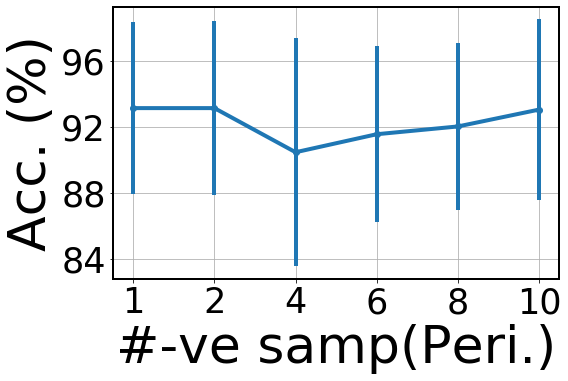}
    \caption{}
    \label{fig:EmbDim}
  \end{subfigure}
  \begin{subfigure}[b]{0.19\linewidth}
    \includegraphics[width=\linewidth]{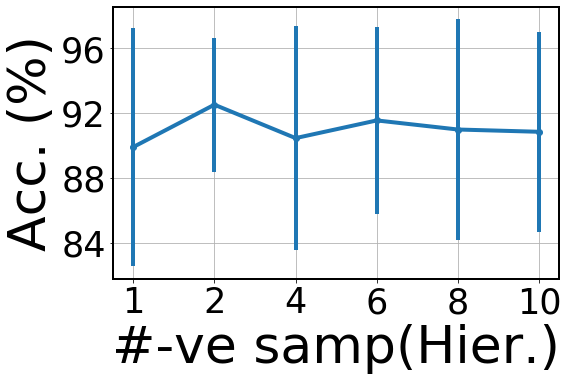}
    \caption{}
    \label{fig:EmbDim}
  \end{subfigure}
  \begin{subfigure}[b]{0.19\linewidth}
    \includegraphics[width=\linewidth]{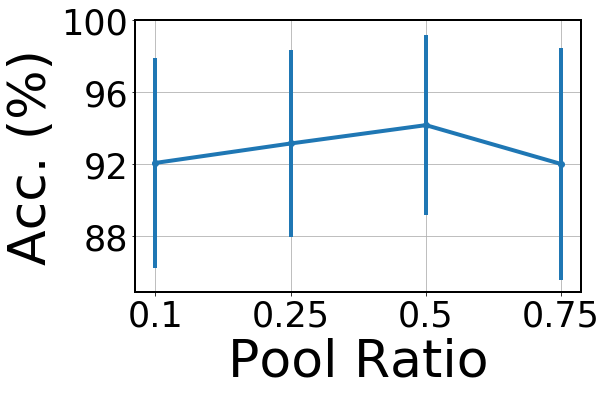}
    \caption{}
    \label{fig:EmbDim}
  \end{subfigure}
    \caption{Sensitivity analysis of GraPHmax for graph classification on MUTAG with respect to different hyperparameters: (a) Number of level graphs, (b) Embedding dimension $K$, (c) Number of negative samples for each positive sample in the periphery discriminator, (d) Number of negative samples for each positive sample in the hierarchical discriminator and (e) Pooling ratio.}
  \label{fig:sensityMutag}

\end{figure*} 

\begin{figure*}[h!]
  \centering
  \begin{subfigure}[b]{0.19\linewidth}
    \includegraphics[width=\linewidth]{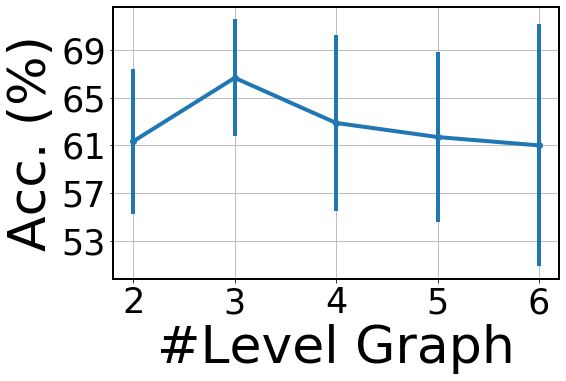}
    \caption{}
    \label{fig:EmbDim}
  \end{subfigure}
  \begin{subfigure}[b]{0.19\linewidth}
    \includegraphics[width=\linewidth]{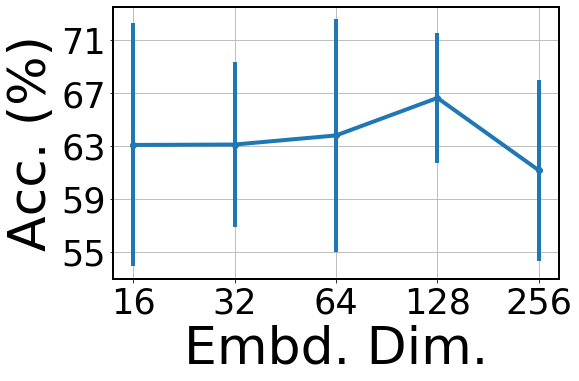}
    \caption{}
    \label{fig:EmbDim}
  \end{subfigure}
  \begin{subfigure}[b]{0.19\linewidth}
    \includegraphics[width=\linewidth]{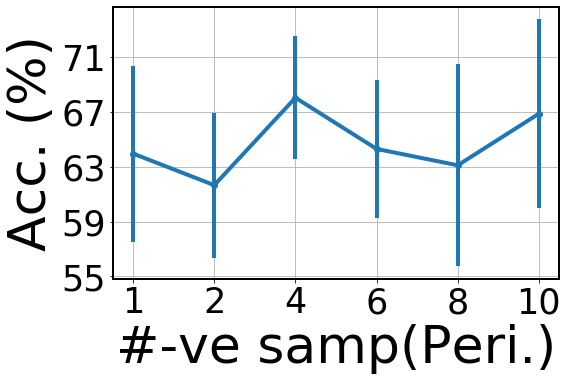}
    \caption{}
    \label{fig:EmbDim}
  \end{subfigure}
  \begin{subfigure}[b]{0.19\linewidth}
    \includegraphics[width=\linewidth]{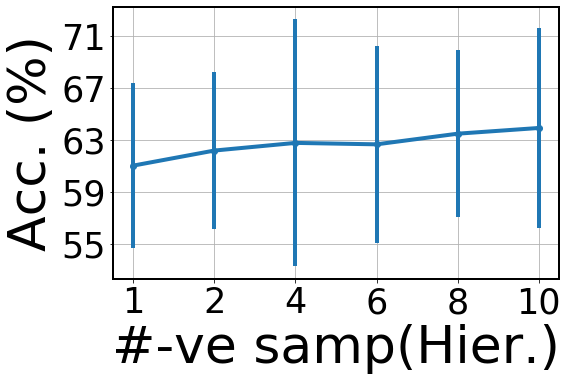}
    \caption{}
    \label{fig:EmbDim}
  \end{subfigure}
  \begin{subfigure}[b]{0.19\linewidth}
    \includegraphics[width=\linewidth]{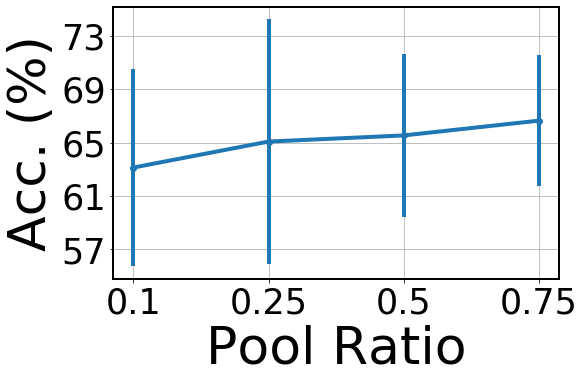}
    \caption{}
    \label{fig:EmbDim}
  \end{subfigure}

  \caption{Sensitivity analysis of GraPHmax for graph classification on PTC for the same set of hyperparameters as in Figure \ref{fig:sensityMutag}}
  \label{fig:sensityPTC}

\end{figure*} 

\subsection{Results on Graph Classification}\label{sec:classi}
Table \ref{tab:graphClassi} presents the results of graph classification. Results of all the baseline algorithms are collected from their respective papers and from the recent literature \cite{NIPS2019_Provably}. Thus, we avoid any degradation of the classification accuracy of the baseline algorithms due to insufficient hyperparameter tuning. It can be observed that our proposed algorithm GraPHmax, though unsupervised in nature, is able to improve the state-of-the-art performance (including the recently proposed supervised GNNs) on MUTAG and PTC.
Besides, it remains to be the \textit{best performing algorithm among the set of unsupervised graph embedding techniques} (the last 10 algorithms in Table \ref{tab:graphClassi}) \textit{on all the datasets}. We can also see a significant gap in the performance of GraPHmax with all the four variants, which clearly convey the importance of all the components (i.e., the use of periphery representation via the first two variants, periphery information maximization via GraPHmax-P and hierarchical information maximization via GraPHmax-H) used in GraPHmax.

\subsection{Graph Clustering}\label{sec:clus}
Graph clustering is an unsupervised task on a set of graphs. We use k-means algorithm (with $k$=\#unique labels) on the vector representations of the graphs obtained from our algorithm, and its four variants. We also consider DGI and InfoGraph as they are recently proposed unsupervised GNNs. We follow the same hyperparameter tuning strategy that was adopted for GraPHmax, for these two baselines. Results show that GraPHmax is able to improve the performance on MUTAG and IMDB-M datasets. DGI performs the best on PROTEINS with all the other algorithms following closely.

\subsection{Graph Visualization}\label{sec:viz}
Graph visualization is the task to plot the vector representations of the graphs in a 2D plane. We use t-SNE \cite{vanDerMaaten2008} on the representations produced by GraPHmax and its variants. For the interest of space, we show the visualization results only on MUTAG dataset. Again the superior performance of GraPHmax over its variants show the usefulness of individual components of GraPHmax.

\subsection{Hyperparameter Sensitivity Analysis}\label{sec:sensitivity}
Here, we check the sensitivity in the performance of GraPHmax for graph classification with respect to a set of important hyperparameters. We have shown the results on MUTAG and PTC datasets in Figures \ref{fig:sensityMutag} and \ref{fig:sensityPTC} respectively. First, the experiment on the number of level graphs in the hierarchical representation (Section \ref{sec:hierar}) of GraPHmax shows that the optimal performance can be achieved when number of levels is 3 (input graph, one intermediate level graph and the final graph containing only one node). This is reasonable as the size of individual graphs in both the datasets are small, so it is less likely to have a longer latent hierarchy in them. Please note, when the number of levels is 2, the hierarchical structure boils down to a flat structure and thus the results degrade. With increasing graph embedding dimension, performance more or less increases on both the datasets, and drops at the end for PTC. We can see the performance of GraPHmax is stable with respect to other hyperparameters on both datasets, except with the number of negative samples used in the periphery discriminator ($n_P$ from Section \ref{sec:pinfomax}) on PTC for which $n_P=4$ produces the best result.

\section{Discussions}\label{sec:disc}
Graph representation through unsupervised GNN is a less studied area in the important domain of graph representation learning. In this paper, we propose a novel unsupervised GNN, referred as GraPHmax, which uses the principle of mutual information maximization through periphery and hierarchical representations of a graph. Thorough experimentation and detailed analysis helps us to understand the importance of individual components of GraPHmax and also improve the state-pf-the-art for multiple graph level downstream tasks on several datasets.

\bibliographystyle{ACM-Reference-Format}
\bibliography{GraPHmax}


\pagebreak

\appendix

\section{Supplementary Material on Reproducibility}\label{sec:supp}
In this section, we explain the values of important hyperparameters of our proposed solution GraPHmax. Hyperparameters for Periphery representation and the optimal values are detailed in table \ref{tab:perirep} for all the datasets. \#layers (Intermediate and output) and \#units (Intermediate GIN layers) are the number of GIN layers including the output layer and the number of units in each layer (excluding the output layer) . Another important hyperparameter is the patience value i.e. for how many more epochs to train the model (before early stop) after the loss starts increasing or becomes constant. Hyperparameters of this part of GraPHmax are tuned based on the unsupervised loss, as discussed in Section \ref{sec:infomax}.

Another set of hyperparameters are for Hierarchical Representation part as reported in table \ref{tab:Hierrep} with the optimal values for each dataset. The output of hierarchical representation is fed to periphery and hierarchical discriminators to maximize mutual information. We use logistic regression classifiers for both periphery and hierarchical discriminators.
To get the optimal values of the hyperparameters in this case, we varied the number of level graphs (including the input graph) in \{3,4,5\} and pooling ratio (defined as $\gamma = \frac{N_{r+1}}{N_r}$, $\forall r \leq R-1$) in \{0.01,0.1,0.2,0.25,0.5,0.75\}. Furthermore, to play with the number of negative samples (Periphery or Hierarchical) we fix range to [1,10]. Patience parameter has the same meaning as in Periphery part. These hyperparameters are also tuned based on the combined unsupervised loss of Periphery and Hierarchical discriminators. Both these (Periphery and Hierarchical) models  are trained separately to output the final Graph Representations.

Please note, for hyperparameter sensitivity analysis in Section \ref{sec:sensitivity} of the paper, when we vary one particular hyperparameter on a dataset, values of other hyperparameters are fixed to their default values as shown in Tables \ref{tab:perirep} and \ref{tab:Hierrep} respectively. Source code of GraPHmax is available at \url{https://bit.ly/3bDxbmY} to ease the reproducibility of the results.



\begin{table*}[!htbp]
\centering
\begin{tabular}{*4c}
	\toprule
	\sffamily{Algorithms} & \footnotesize \#layers (Intermediate and output) & \#units (Intermediate GIN layers) &Patience (Early Stopping) \\
    \hline
	\midrule 
    MUTAG&2&128&30\\
    PTC &3&128$\times$128&30\\
    PROTEINS&3&128$\times$128&40\\
    NCI1&2&128&30\\
    NCI109&2&128&30\\
    IMDB-BINARY&3&128$\times$128&30\\
    IMDB-MULTI&3&128$\times$64&30\\
    
    \bottomrule
\end{tabular}
\caption{This table depicts the architecture details for Periphery Representation i.e. GIN encoder for different datasets. GIN encoder has input layer followed by multiple GIN layers and finally an output layer. Dimension of the last GIN layer (Embedding layer) is same as the input dimension. Besides, dimension of intermediate layers depends on the datasets as mentioned in the table. Encoder is trained for maximum of 2000 epochs with patience parameter (for early stopping) varying according to the datasets. Also, Learning rate is set to 0.001 for all the datasets.}
\label{tab:perirep}
\end{table*}

\begin{table*}[!htbp]
\centering
\begin{tabular}{*7c}
	\toprule
	\sffamily{Algorithms} & \footnotesize \#Level Graphs&Pool Ratio& Embd. Dim.& \#-ve samples(Peri.)&\#-ve samples(Hier.)&Patience (Early Stopping)  \\
    \hline
	\midrule
    MUTAG&3&128&0.5&4&4&1000\\
    PTC &3&128&0.75&4&5&1000\\
    PROTEINS&3&128&0.01&4&4&950\\
    NCI1&3&128&0.5&8&5&650\\
    NCI109&3&128&0.5&8&5&650\\
    IMDB-BINARY&3&128&0.25&4&4&850\\
    IMDB-MULTI&3&128&0.25&5&5&1000\\
 
    \bottomrule
\end{tabular}
\caption{Detailed architecture for Hierarchical Representation. For unsupervised training, maximum number of epochs is set to 3000 with patience parameter (for early stopping) varying according to the datasets. Learning rate is set to 0.001 for all datasets.}
\label{tab:Hierrep}
\end{table*}

\end{document}